\documentclass{svproc}
\usepackage{url}

\usepackage[nolist]{acronym}
\usepackage{amsmath}
\usepackage{bm}
\usepackage{booktabs}
\usepackage{color}
\usepackage{cite}
\usepackage{graphicx}
\usepackage{multirow}
\usepackage{tabularx}
\usepackage{siunitx}
\usepackage{url}

\usepackage{enumitem}

\usepackage[normalem]{ulem}

\newacro{ai}[AI]{Artificial Intelligence}
\newacro{rom}[RoM]{Range of Motion}
\newacro{ml}[ML]{Machine Learning}
\newacro{vr}[VR]{Virtual Reality}
\newacro{emg}[EMG]{Electromyography}
\newacro{lr}[LR]{Linear Regressor}
\newacro{mlp}[MLP]{Multi-Layer Perceptron}
\newacro{svr}[SVR]{Support Vector Regression}
\newacro{rf}[RF]{Random Forest}
\newacro{lstm}[LSTM]{Long Short-Term Memory}
\newacro{rmse}[RMSE]{Root Mean Squared Error}

\mainmatter              
\title{Learning Hand State Estimation\\for a Light Exoskeleton}
\titlerunning{Learning Hand State Estimation for a Light Exoskeleton}  
\author{Gabriele Abbate\inst{1} \and Alessandro Giusti\inst{1} \and Luca Randazzo\inst{2} \and Antonio Paolillo\inst{1}}
\authorrunning{Gabriele Abbate et al.} 
\tocauthor{Gabriele Abbate, Alessandro Giusti, Luca Randazzo, and Antonio Paolillo}
\institute{Dalle Molle Institute for Artificial Intelligence (IDSIA), USI-SUPSI, 
\\ Lugano, Switzerland 
\\ \email{name.surname@idsia.ch}
\and
Emovo Care, EPFL Innovation Park, Lausanne, Switzerland
\\ \email{luca@emovocare.com}}
\begin{document}
\maketitle              

\begin{abstract}
We propose a machine learning-based estimator of the hand state for rehabilitation purposes, using light exoskeletons.
These devices are easy to use and useful for delivering domestic and frequent therapies. 
We build a supervised approach using information from the muscular activity of the forearm and the motion of the exoskeleton to reconstruct the hand's opening degree and compliance level. 
Such information can be used to evaluate the therapy progress and develop adaptive control behaviors.
Our approach is validated with a real light exoskeleton. 
The experiments demonstrate good predictive performance of our approach when trained on data coming from a single user and tested on the same user, even across different sessions.
This generalization capability makes our system promising for practical use in real rehabilitation.
\end{abstract}
%
\section{Introduction}

Stroke is one of the main causes of disability~\cite{Deuschl2020}, resulting in severe hand functionalities limitation~\cite{Kwakkel473}, or long-lasting hand impairments~\cite{Kwakkel2003}.
Given the importance of manual operations in everyday life, rehabilitation procedures are given a crucial role~\cite{Raghavan2007, hunter_crome_2002}.
Cutting-edge technology, like \ac{vr}~\cite{Heinrich2022, Levin2015} and robotics~\cite{Abbate:scirep:2023}, can assist standard rehabilitation to achieve better outcomes.
%
In this context, it is expected that \ac{ai} can bring great benefits, especially in solving perception challenges that are otherwise difficult to tackle with standard devices.
Indeed, \ac{ml}, and \ac{ai} in general, have been proven to be effective in tackling complex perception tasks in robotics, 
e.g. for complex localization problems~\cite{Nava:ral:2021,Nava:ral:2022} or human-robot interaction purposes~\cite{Abbate:ras:2024,Arreghini:icra:2024}.
\ac{ai} finds application also in the domain of medical robotics~\cite{Yip2023}.

Hand exoskeletons are very useful in rehabilitation (see e.g.~\cite{Kabir2022, Veerbeek2017}). 
Equipping them with an onboard and light perception module would enable even better therapy outcomes for several reasons.
First, a perception system that does not require complex infrastructure or additional heavy devices confers versatility and lightness on the exoskeleton.
These aspects are particularly important for favoring ease of use, and domestic and frequent utilization, a key factor for a successful therapy~\cite{Krakauer2006, Dobkin2004}. 
Secondly, an online and robust perception of the patients' state would allow adaptive closed-loop exoskeleton control 
%
with a positive impact on the therapy, as tuning the therapy to the patient's state positively affects motor learning~\cite{Choi2008, Metzger2014}. 
Finally, and very importantly, a perception system would permit 
the measure of the patients' sensorimotor deficit, crucial for delivering optimal rehabilitation~\cite{Lambercy2012}.
At the moment, the therapy verification is manually performed by clinicians and 
%
suffer from low reliability and standardization, often affected by the therapist's perception of the patient's abilities~\cite{Gordon1987}. 

%
Robotics could offer accurate and objective assessments of function and impairment~\cite{Maciejasz2014, Lambercy2012}.
\begin{figure}[t]
    \vspace{2mm}
    \centering    \frame{\includegraphics[angle=-90,trim={37.5cm 0 3cm 0},clip,width=0.8\columnwidth]{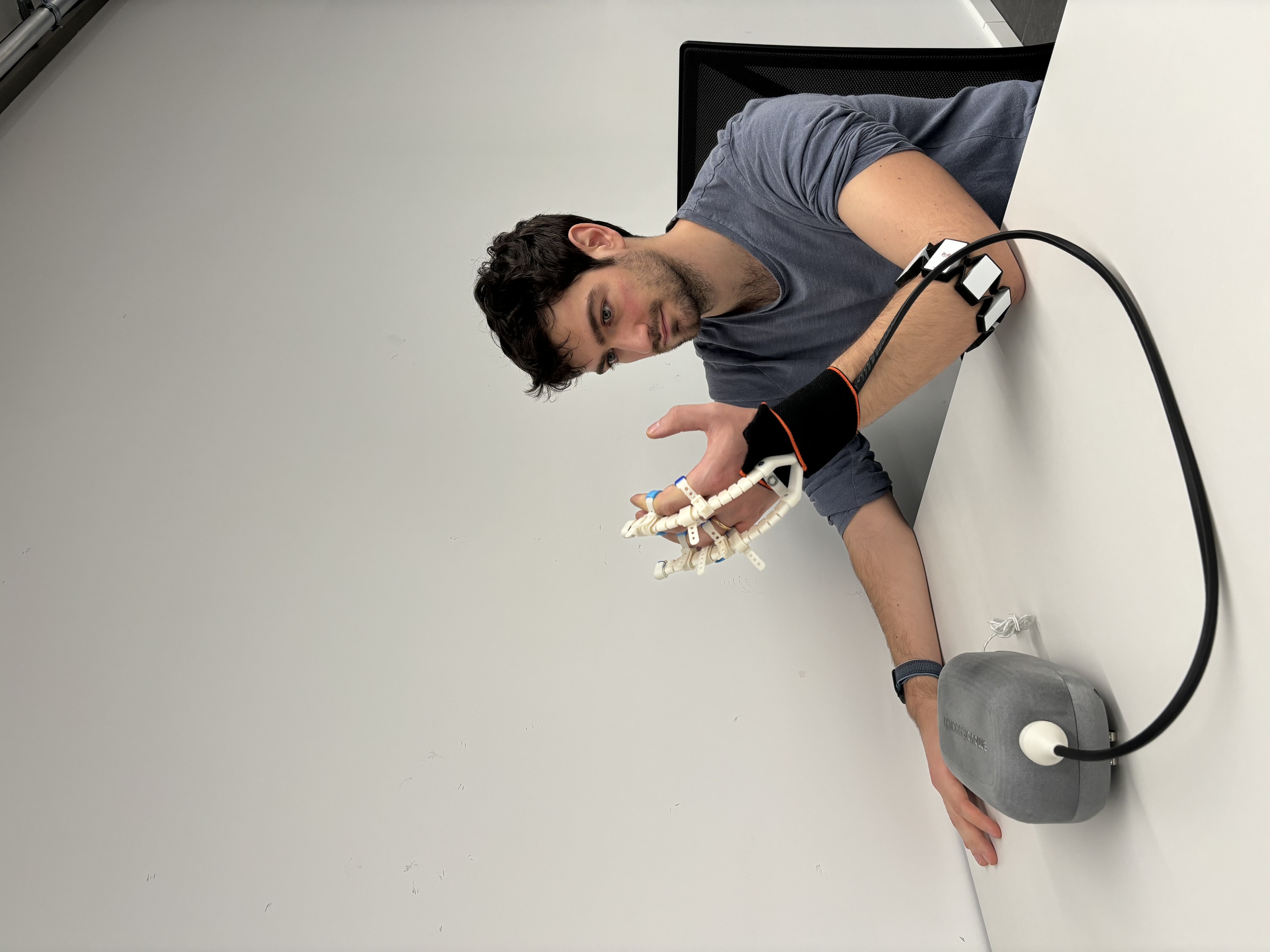}}
    \caption{Light exoskeletons are promising and powerful tools for effective rehabilitation therapies. The challenge addressed in this work is to provide this kind of device, having little sensory equipment, with advanced perception systems to online measure the patient's state and the therapy progress.}
    \label{fig:intro}
\end{figure}
For example, robotic devices are deployed to evaluate 
proprioception and haptic perception~\cite{Metzger2014}, and finger \ac{rom}~\cite{Su:icra:2024, Wang2021}.
Fingers' \ac{rom} are predictors of rehabilitation outcomes~\cite{Marek2023} and could be measured using hand trackers.
These systems need external infrastructures, be it cameras~\cite{Lugaresi2019} or other specialized hardware, such as gloves~\cite{Almeida2019}.
%
However, vision-based approaches have limited tracking area, as they are constrained by the camera field of view, and are susceptible to external disturbances such as light changing and occlusion. 
While gloves are robust in that sense, they may not be suitable to be used together with exoskeletons. 
%
%
%
Also, an under-actuated exoskeleton has been proposed to track the user's fingers \ac{rom}~\cite{Su:icra:2024}, but it can not perform rehabilitation, since it only provides force feedback at the fingertips level.
Ultimately, it is desirable to measure the hands \ac{rom}, for further evaluation of the therapy effectiveness, with no heavy infrastructure and complex wearing procedure.
A dry surface \ac{emg} sensor is a compelling alternative, as it is easy to wear and its measure of muscle activation can serve the monitoring of neuromuscular pathologies~\cite{Campanini2020}.

Light hand exoskeletons (as the one shown in Fig.~\ref{fig:intro}) offer interesting possibilities for rehabilitation as they are: easy to wear and simple to use for both patient and therapist; flexible and compliant so that they easily adapt to the needs of stroke patients (who might have stiffness in the hand); relatively inexpensive, which favors their large usage.
On the other side, their mechanism is simple and lacks rich sensory equipment.
Thus, building an advanced perception system able to measure the patient's state and the therapy progress requires tackling technological and research challenges.

This work proposes to provide light hand exoskeletons with the perception of the patient's hand state to improve the current rehabilitation setups and unlock possibilities in this domain.
Our perception module (Sec.~\ref{sec:approach}) is built on the measurement of the forearm muscle activity, obtained with \ac{emg} sensors, and the exoskeleton motion. 
%
%
A supervised 
approach 
learns the actual behavior of the hand wearing the exoskeleton.
The experimental setup, the data collection procedure, and other implementation details are described in Sec.~\ref{sec:experimental_setup}.
Results are presented in Sec.~\ref{sec:results}, whereas Sec.~\ref{sec:conclusions} concludes the paper with final remarks.

\section{Problem formulation and proposed approach}\label{sec:approach}

We address the problem of building a perception module for light hand exoskeletons with little sensory equipment.
The aim is to reconstruct the state of the patient's hand, who wears a hand exoskeleton and an \ac{emg} sensor, in a rehabilitation setup.
In detail, we tackle the challenge of estimating the opening degree of the patient's hand and its level of compliance.
The first can be used to estimate the patient's \ac{rom}, for the evaluation of the therapy progress; the second can be used to adapt the action of the exoskeleton to the patient's current state.
We denote the opening degree with a continuous variable $y_o \in [0,\pi/2]$, where $y_o=0$ means `open hand' and $y_o=\pi/2$ is for `closed'.
The hand's compliance level is defined with the dimensionless variable $y_c \in [-1, 1]$ ranging from $y_c=-1$ (for stiff hand) to $y_c=1$ (compliant hand), passing through $y_c=0$ (neutral).
The quantities $y_o$ and $y_c$ compose the target variable $\bm{y}$ to estimate:
\begin{equation}
    \bm{y} = \left( y_o, y_c \right)^\top.
    \label{eq:target}
\end{equation}

We pursue our perception objective without resorting to any external hand-tracking systems.
%
%
Instead, we use the information coming from the motion of the exoskeleton (such as the position of its motors) collected in a feature set called $\bm{f}_\text{exo}$, and the corresponding muscular activity of the patient detected through the \ac{emg} sensor collected in another feature set, $\bm{f}_\text{emg}$.
Individually, these features are poorly informative of the real hand motion.  
On one side, the exoskeleton motors do not take into account non-modeled effects such as the flexibility of the tendons or the backlash with the patient's fingers.
On the other, the noisy \ac{emg} signals do not allow accurate hand motion tracking.
We claim, instead, that fusing these pieces of information can build an accurate estimation of the hand behavior.
Therefore, the union of the exoskeleton and \ac{emg} feedback represents the input feature set: 
\begin{equation}
    \bm{f} = \left( \bm{f}_\text{exo}^\top, \bm{f}_\text{emg}^\top \right)^\top.
    \label{eq:features}
\end{equation}

We build a supervised \ac{ml} approach, consisting of a model $\bm{m}$ that estimates the target from  the input feature set:
\begin{equation}
	\hat{\bm{y}} = \bm{m}\left(\bm{f}|{\cal D}\right)
	\label{eq:model}
\end{equation}
where the hat over the variable indicates its estimated value.
%
%
The training dataset ${\cal D}$ consists of a set of sequences labeled with the ground truth values of $\bm{y}$:
\begin{equation}
    {\cal D} = \left\{ \bm{f}_{i,j}, \bm{y}_{i,j} \right\}_{i=1,j=1}^{N_j,S}.
    \label{eq:dataset}
\end{equation}
%
%
The subscripts $i$ and $j$ denote the $i$-th sample of the $j$-th sequence, respectively; $N_j$ is the number of samples of the $j$-th sequence; $S$ is the number of sequences.
%
%





\section{Experimental setup}\label{sec:experimental_setup}
\subsection{Hardware and sensory equipment}

\begin{figure}[t]
\vspace{2mm}
    \centering
    \begin{minipage}{0.41\columnwidth}      \frame{\includegraphics[angle=-90,trim={33cm 0 14cm 0},clip,width=\linewidth]{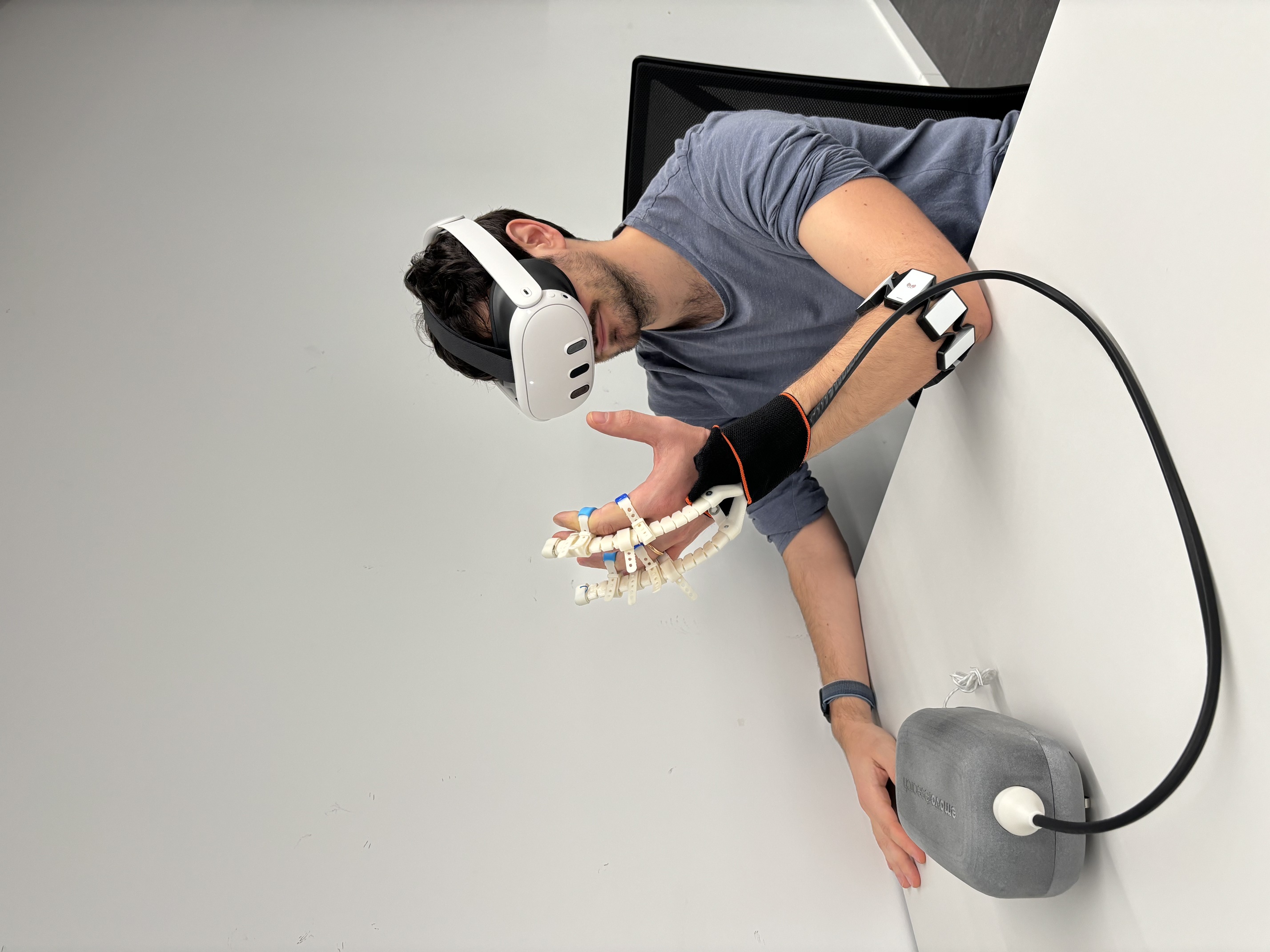}}
    \\[11pt]
    \frame{\includegraphics[trim={6.5cm 2cm 0cm 21cm},clip,width=\linewidth]{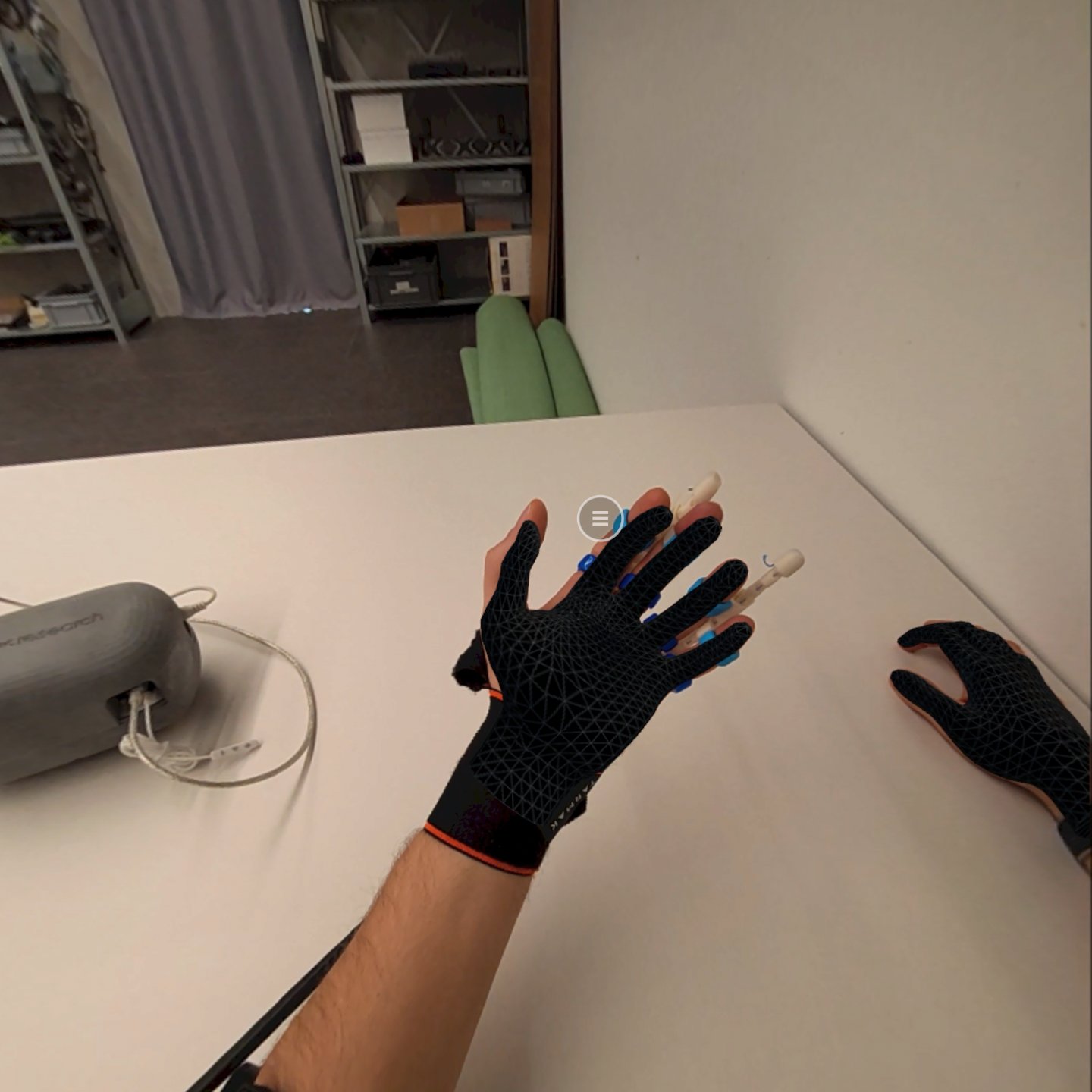}}    
    \end{minipage}
    ~
    \begin{minipage}{0.3\columnwidth}
    \frame{\includegraphics[angle=0,trim={2.5cm 6.6cm 17.6cm 0cm},clip,width=\linewidth]{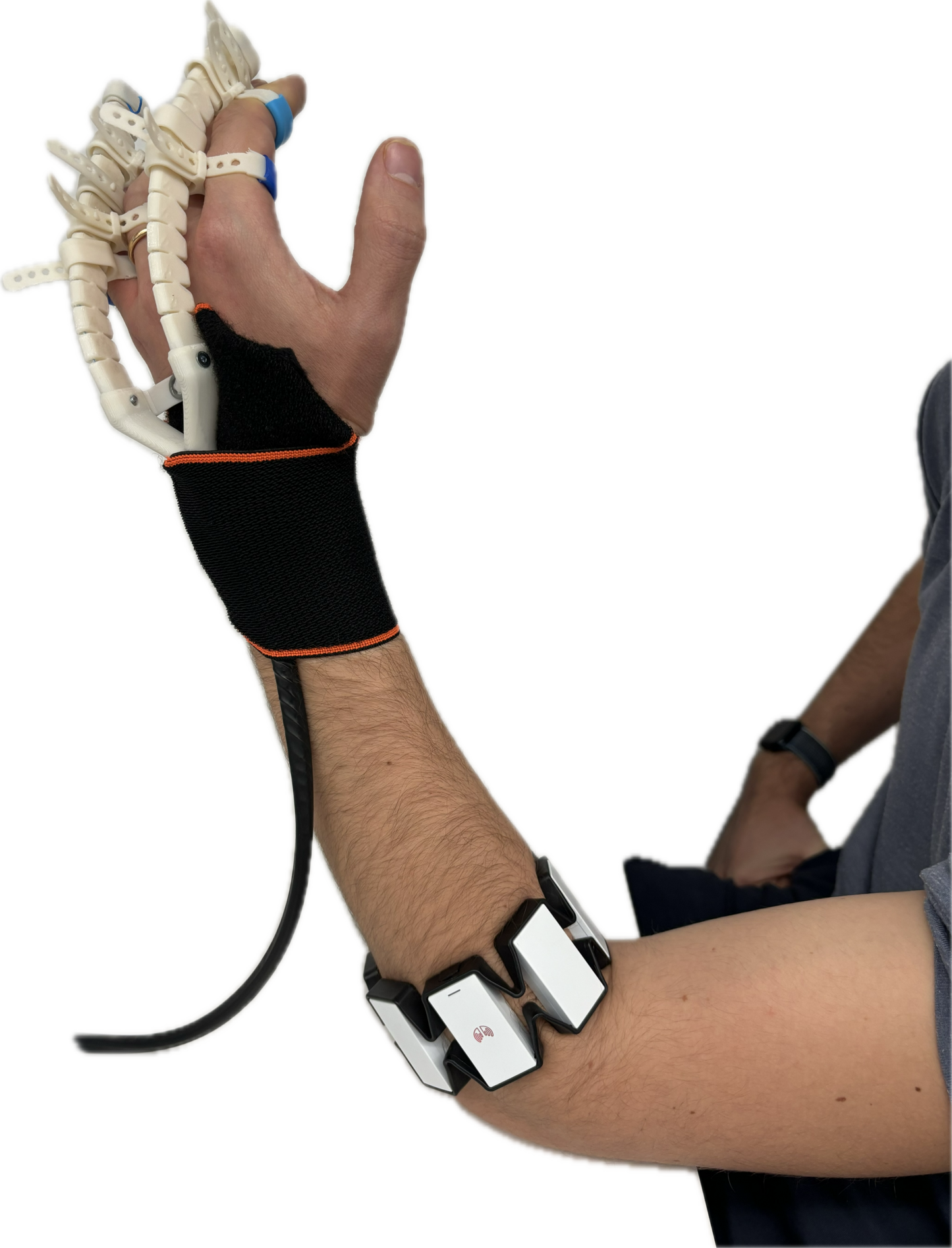}}    
    \end{minipage}
    \caption{Data collection setup: a user wears the exoskeleton and the \ac{emg} sensor at the left hand and forearm, respectively, and an augmented reality headset (top left); detailed view on the exoskeleton and the \ac{emg} sensor (right); view of the augmented reality scene (bottom left).}
    \label{fig:setup}
\end{figure}



We use the hand exoskeleton built by Emovo Care\footnote{\url{https://emovocare.com}} for specific rehabilitation purposes. 
The mechanical structure of this exoskeleton consists of two soft, flexible tendons that are worn coupling the index with the middle finger, and the ring finger with the pinkie (see Fig.~\ref{fig:setup}).
A single motor actuates both tendons simultaneously, assisting simple opening and closing hand motions. 
Such a light and simple structure is easy to use for both patients and therapists.
In particular, the soft tendons allow passive compliance, so that the exoskeleton can simply adapt to the patient's hand posture. 
The exoskeleton is also simple in terms of sensory equipment: an encoder streams the motor position at \SI{20}{\hertz}; the current applied to the motor is also read at the same frequency. 
These measurements compose the vector $\bm{f}_\text{exo}$ defined in Sec.~\ref{sec:approach}.
The device is also equipped with Wi-Fi and serial USB communication interfaces to send commands to the motor and read its state. 
More details are in the description of an early device prototype~\cite{Randazzo:ral:2018}.

To measure the muscular activity of the hand, we use the Myo armband, an $8$-channel dry surface \ac{emg} sensor widely used in combination with arm prostheses~\cite{Visconti2018}. 
The device (visible in Fig.~\ref{fig:setup}) consists of $8$ pads that, in contact with the muscles of the forearm, detect the activity of the user's hand opening motion.
The Myo system has a marker on one of the pads, allowing different users to consistently wear the sensor at the same position (i.e. with the marker facing up and pointing towards the user's shoulder).
The arm motion activity is decoded in the form of $8$ raw electrical signals, one for each pad, which are further processed by the device's onboard standardization and filtering operations.
Ultimately, the sensor streams $8$ filtered signals via a Bluetooth interface at \SI{50}{\hertz}, that fill the vector $\bm{f}_\text{emg}$ defined in Sec.~\ref{sec:approach}.

\subsection{Data collection}

\label{sec:data_collection}
For collecting data, $5$ users ($3$ M, $2$ F, ages $25$--$64$) are asked to wear the Myo sensor and the exoskeleton on their left forearm and hand, respectively (Fig.~\ref{fig:setup}). Participants gave their informed consent to participate and the procedure was approved by the local ethics committee of the University of Applied Sciences and Arts of Southern Switzerland.

To label our data, during the acquisition, we use an external hand-tracking system.
%
%
We opt for the Meta Quest 3 Headset\footnote{\url{https://www.meta.com/ch/en/quest/quest-3/}} as its built-in capabilities allow precise and high frequency (\SI{90}{\hertz}) tracking of 
hand joints.  
%
We define the hand's opening degree as the average values of the pitch angles of each finger joint, excluding the thumb as it is not actuated by the exoskeleton. 
This quantity, as provided by the hand tracker, represents the ground truth 
of the target $y_o$ defined in Sec.~\ref{sec:approach}. 
Also, we leverage the augmented reality features of the headset to make data acquisition comfortable for the users.
As shown in Fig.~\ref{fig:setup}, users can 
see their real surroundings while a virtual hand (which is a visualization of the tracking output) is superimposed on the real one. 

During the acquisition, the exoskeleton performs cyclic opening and closing motions, in periods of $10$~s.
Notably, this periodic motion, used to train our perception model, does not imply that the exoskeleton has to move in the same way during the deployment. 
%
Each data sequence lasts $60$~s (i.e., consists of $3$ open-close cycles); users are asked to maintain the same hand compliance for each sequence.
In particular, we designed the following three acquisition modalities to emulate the different levels of participation of a possible stroke patient:
\begin{description}[font=\normalfont]
    \item[\emph{Passive}:] the users do not perform any movement and let the exoskeleton open and close their hand; 
    \item[\emph{Helping}:] the users actively perform the movement induced by the exoskeleton; 
    \item[\emph{Opposing}:] the users actively resist the movement started by the exoskeleton.
\end{description}
%
%
The hand compliance value, known in advance, 
is manually assigned to the data sequences, thus defining the ground truth of the variable $y_c$ defined in Sec.~\ref{sec:approach}.
%
We assign $y_c = 1$ (compliant hand) to the sequences acquired during the Helping modality, $y_c = 0$ (neutral) to the Passive data, and $y_c =-1$ (stiff) for Opposing.
We collect $9$ sequences ($3$ for each modality) for the $5$ users, totaling $45$ sequences. 

\subsection{Model architectures and evaluation metrics}\label{sec:models}

Data from the three devices (\ac{emg} sensor, exoskeleton, and hand tracker) are acquired in real time, stored, and synchronized on an external laptop. 
%
%
Such data, collected together in the dataset ${\cal D}$, are then used to train and evaluate our perception model.
We compare different architectures: \ac{lr}, \ac{mlp}, \ac{svr}, and a stateful \ac{lstm}-based regressor. 
The stateless models are implemented with scikit-learn library\footnote{\url{https://scikit-learn.org/}} using default parameters, except for the \ac{mlp} which consists of $2$ hidden layers with $100$ neurons each.
The stateful \ac{lstm}-based model, instead, is implemented in PyTorch\footnote{\url{https://pytorch.org}} and is composed of $2$ \ac{lstm} cells with a $32$-dimensional hidden state each.
Both \ac{mlp} and \ac{lstm} models have
approximately $12000$ trainable parameters.
As a baseline, we additionally report a \emph{Dummy} regression model that always returns the average of the ground truth from the training data.

The different models are evaluated using the \ac{rmse} and the coefficient of determination, denoted with $R^2$.
The former measures the average difference between model prediction and ground truth values; an ideal regression provides RMSE equal to $0$.
The latter is a dimensionless value measuring the amount of variance in the target variable that is explained by the model. 
An ideal regression model yields $R^2=1$, while a naive regression that estimates the mean of the target variable yields $R^2=0$.

%
\section{Results}\label{sec:results}

\begin{figure}[!t]
\centering
\vspace{2mm}
    \centering
    \begin{minipage}{0.75\linewidth}%
    \includegraphics[width=\columnwidth]{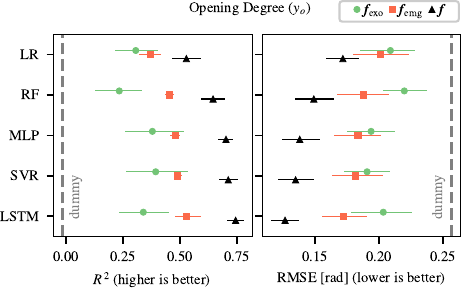}
    \\[15pt]
\includegraphics[width=\columnwidth]{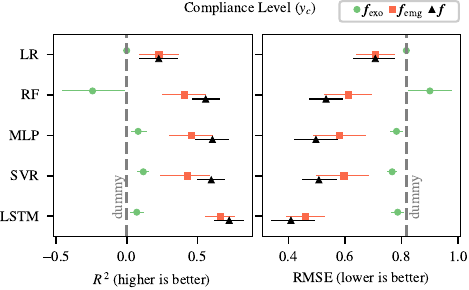}    
    \end{minipage}
    \caption{Prediction performance of the different regression architectures ($y$-axis) trained with different feature sets (colors): 
    $R^2$ (left, higher is better) and \ac{rmse} (right, lower is better) averaged among users and reported 
    for $y_{o}$ (top) and $y_{c}$ (bottom). Error bars denote $80\%$ confidence interval. 
    }
    \label{fig:kfold_results}
\end{figure}

\subsection{Prediction performance}

We evaluate the prediction performance of the different regression architectures (introduced in Sec.~\ref{sec:models}) and the impact of different feature sets.
To this extent, we train each architecture with the full feature set defined in Sec.~\ref{sec:approach} (i.e., $\bm{f}$) and with the individual feature sets coming from the exoskeleton and the Myo sensor ($\bm{f}_\text{exo}$ and $\bm{f}_\text{emg}$).
This comparison serves to confirm our claim that data of the exoskeleton and the \ac{emg} alone are not informative enough; instead, fusing them is beneficial. 
We also show that our approach is capable of learning to estimate the hand state of a given user using only a small amount of data (i.e. $9$ minutes) recorded from the same user.
%
%
To this end, we perform $3$-fold cross-validation independently for each user.
Each fold is built to have $3$ validation sequences, one for each hand compliance acquisition modality, as defined in Sec.~\ref{sec:data_collection}. Also, each train set contains $6$ training sequences, $2$ for each acquisition modality.
For each combination of user, architecture, and set of features we train 3 models, each corresponding to one validation fold. 
For each combination, we get predictions for the sequences in a fold using the corresponding model, then we compute the metrics across the $3$ folds considered together.

Figure~\ref{fig:kfold_results} reports both the evaluation metrics (columns) for each target variable (rows), model architecture (spread on the $y$-axis of the plots), and set of features (reported in different colors), averaged among all users.
%
\ac{lstm} consistently outperforms 
the others (i.e. 
higher $R^2$ and lower \ac{rmse} values). 
These values also show that the prediction capabilities of models trained only with $\bm{f}_\text{emg}$ or $\bm{f}_\text{exo}$ 
are lower than those trained with $\bm{f}$.
In particular, \ac{lstm} trained with $\bm{f}$ yield on average the highest $R^2$ 
and the lowest \ac{rmse}, 
i.e., $R^2 = 0.74$ and \ac{rmse} equal to $0.12$ for the opening degree; $R^2 = 0.72$ and \ac{rmse} equal to $0.40$ for the compliance level.

\begin{figure*}[!t]
    \centering    
    \includegraphics[width=0.975\textwidth]{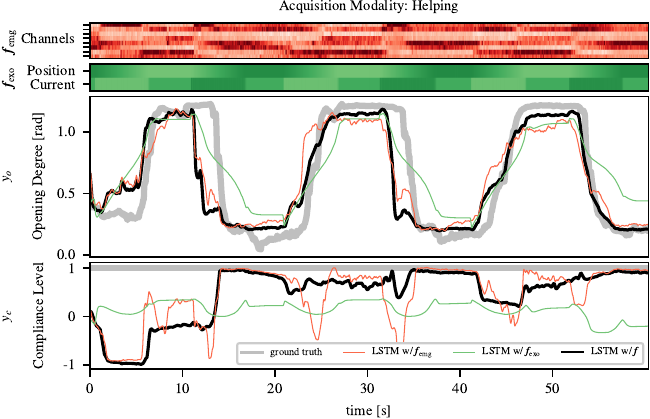}
    \caption{
    A sequence of data acquired by a single user for the helping acquisition modality and the corresponding target prediction.
    }
    \label{fig:single_user_helping}
\end{figure*}

\begin{figure*}[!t]
    \centering    
    \includegraphics[width=0.975\textwidth]{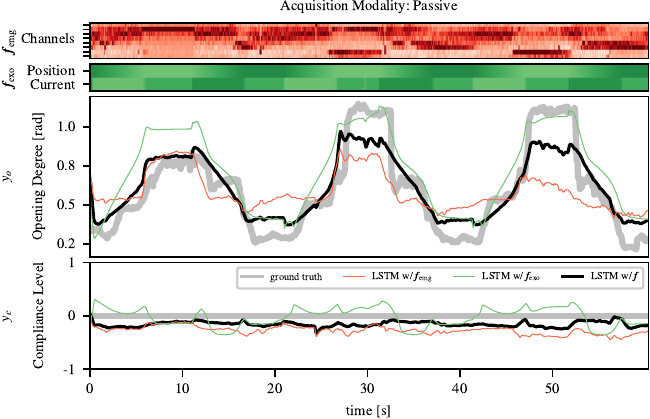}
    \caption{
    A sequence of data acquired by a single user for the passive acquisition modality and the corresponding target prediction.
    }
    \label{fig:single_user_passive}
\end{figure*}

\begin{figure*}[!t]
    \centering    
    \includegraphics[width=0.975\textwidth]{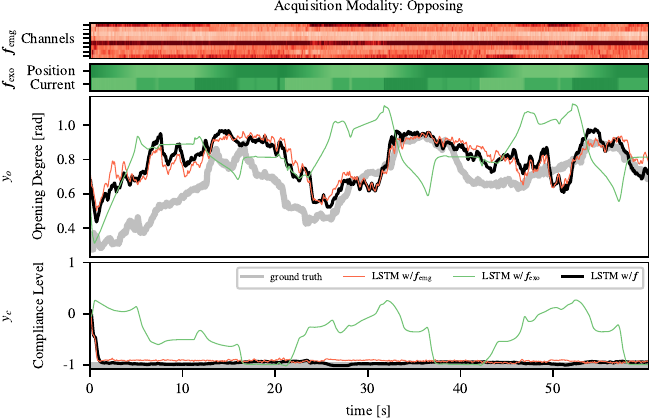}
    \caption{
    A sequence of data acquired by a single user for the opposing acquisition modality and the corresponding target prediction. 
    }
    \label{fig:single_user_opposing}
\end{figure*}

These performances are further highlighted in Figs.~\ref{fig:single_user_helping}--\ref{fig:single_user_opposing}. 
Here we select a single user and the corresponding \ac{lstm} models trained with the 3-fold cross-validation on their data. 
As explained, we have 3 models trained in different folds for each feature set. 
We average the prediction of these 3 models over a test set collected on the same user during another session, and that is not used during training. 
This test set is composed of 3 sequences, one for each 
acquisition modality, which are reported in each figure.
Looking at the second row in each plot, one can notice that the $\bm{f}_\text{exo}$ features bring too little information since they display only small changes across the different sequences. 
%
This is due to the soft and compliant nature of the exoskeleton, which introduces flexibility and backlash to the actuation systems.
As a result, the motor motion does not match the one of the user's hand.
In particular, the motor actuating the exoskeleton can reach the target position even when the user's hand is not compliant at all, without recording significant changes in the $\bm{f}_\text{exo}$ features.
Consequently, the model based on $\bm{f}_\text{exo}$ (green lines in the plots) shows low predictive capabilities, particularly for the hand compliance target (last rows in Figs.~\ref{fig:single_user_helping}--\ref{fig:single_user_opposing}) or the sequence with low hand compliance. 
On the contrary, $\bm{f}_\text{emg}$ features (first rows) clearly show different patterns of activation across different types of sequences resulting 
in a better prediction performance (red lines).
Finally, the model using the full 
set $\bm{f}$ 
performs best: black lines are close to the gray ones, 
i.e., the ground truth.


\subsection{Generalization capabilities}

\subsubsection{Generalization across different sessions} 
\begin{figure}[ht!]
    \centering    
    \includegraphics[width=0.75\columnwidth]{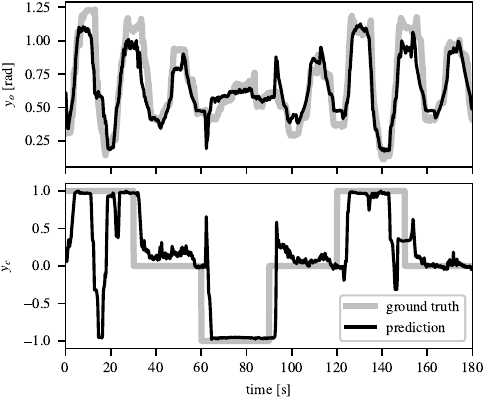}
    \caption{Online experiment: prediction vs ground truth of the target variables.  
    }
    \label{fig:online}
\end{figure}

We verify how well a model trained on one user during a session, performs on data from the same user collected in a different session. On a different day, users have to wear and position the sensors again, potentially affecting the system's performance. 
Figs.~\ref{fig:single_user_helping}--\ref{fig:single_user_opposing} and Fig.~\ref{fig:online} report examples of this scenario.
They demonstrate that models trained with just $9$ minutes of data coming from a single user can generalize well for the same user during a different session.
This aspect is crucial to make the system usable in practice.
%
For example, in a real clinical scenario, a therapist would perform a calibration when training a patient to execute the therapy. In this way, the headset, or an alternative device to generate ground truth, would be needed only on this occasion.
Users could carry on with the therapy on their own needing only the exoskeleton and the \ac{emg} sensor.

\begin{figure}[t]
    \centering    
    \includegraphics[width=0.75\columnwidth]{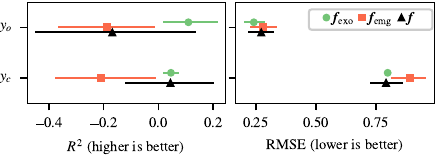}
    \caption{Generalization across users: 
    $R^2$ and \ac{rmse} averaged across users, reported for each target variable and feature set. 
    }
    \label{fig:user_vs_users}
\end{figure}

\subsubsection{Generalization across users} We verify whether a model, trained with all available data, generalizes on data coming from a new user.
To this end, we select the best-performing architecture from the previous analysis (i.e. \ac{lstm}-based) and we deploy a new cross-validation strategy. 
We train a model for each user with data available from all the other users. Then, we evaluate the performance of each model on the corresponding user who was kept out of training.
Figure~\ref{fig:user_vs_users} shows low performance across all users both in terms of $R^2$ and \ac{rmse}. 
This means that our method does not generalize to new users, without specific training.
This can be explained by the fact that $\bm{f}_\text{emg}$ features are strictly dependent on the subject~\cite{Lu2020}, so each user should undergo a calibration procedure before starting to use the system.
This is further marked by the fact that in Fig.~\ref{fig:user_vs_users} the best-performing architecture is the one based on $\bm{f}_\text{exo}$ since these features are less dependent on a specific user. However, as already demonstrated, exoskeleton data alone are not informative enough to obtain a good predictor.
%
%

\subsection{Online experiments}

We execute an online experiment where we train an \ac{lstm} regressor based on $\bm{f}$ features with data coming from a single user. 
The data is collected as explained in Sec.~\ref{sec:data_collection}.
Then, on a different day, the same user is asked to perform again the task as during the data collection, but with a difference: every $30$~s, as suggested by an audio clue, they are asked to switch to a different acquisition modality.
In this way, we record a $180$~s long sequence during which the user loops back and forth between \emph{Helping}, \emph{Passive}, and \emph{Opposing} acquisition modality.
At the same time, the model previously trained is deployed in real-time to predict both target variables.
Figure~\ref{fig:online} shows the prediction compared to the ground truth values.
It qualitatively displays good performance, further confirmed by the $R^2$ and \ac{rmse} values ($0.81$ and $0.12$ for $y_{o}$, $0.55$ and $0.45$ for $y_{c}$).
The experiment execution is recorded in the video available online\footnote{\url{https://youtu.be/6PiVOs6P4JA}}.
Note that the user wears the headset just to record $y_{o}$ ground truth for validation purposes. 
In an application deployment, the headset is not needed (i.e. the setup would be as in Fig.\ref{fig:intro}).


\section{Conclusions}\label{sec:conclusions}
We have presented a machine learning-based approach to estimate the hand state of a user wearing a light exoskeleton for rehabilitation purposes. 
More in detail, our objective is to predict the opening degree of the hand and its level of compliance. 
The former is an indicator of the user's \ac{rom}; the latter is a measure of the user's active participation in the therapy. 
Both are useful to objectively asses a therapy outcome and continuously adapt it to the patient's state.
We have demonstrated that using the exoskeleton alone is not enough to achieve our goal.
Indeed, the soft nature of the device introduces slacks between the hand joints and the exoskeleton fingers, making the data coming from its motor poorly informative of the actual hand state, in particular for low hand compliance. 
%
We proposed to add an EMG sensor to the setup, which can enrich the simple perception capabilities of the exoskeleton with more insightful sensing. 
Our experiments show that fusing the two data sources allows an \ac{lstm}-based regression to obtain a strong prediction performance even when we train with little data on a single user, generalizing well across different sessions of the same user. 
Such capability makes the approach suitable for use in a real-world scenario, even if the model does not generalize well across different users. 
In fact, a patient could easily and quickly calibrate the system at their very first use, using a hand-tracking device with the help of a therapist. 
Then, they could keep up with the therapy on their own, thanks to the portable nature of the proposed system.

The main limitation of the current work is that experimentation is done with healthy users. 
Our promising results suggest that the estimation capabilities of our system can potentially translate to patients with hand injuries; however, tests with actual patients are needed.
Future work will be devoted to the assessment of the tool's usability and acceptability with stroke patients, which strongly impact the success of the therapy. 
%
%
We have asked our participants to simulate different --- well distinguishable --- levels of hand compliance.
However, monitoring 
the hand behavior of actual patients may be more difficult.
Definitely, a study involving patients affected by hand injuries is a clear path for future research.

\section*{Acknowledgement}
This work was supported by Innosuisse - Swiss Innovation Agency, through the project “Virtual Reality and Hand Exoskeleton for Mirror Therapy: a Feasibility Study (VRHEM)” (100.533 IP-ICT).

\bibliographystyle{splncs03}
\bibliography{bibliography}







\end{document}